\documentclass[letterpaper, 10 pt, conference]{ieeeconf}
\IEEEoverridecommandlockouts
\usepackage{amsmath,amsfonts,amssymb}
\usepackage{algorithmic}
\usepackage{algorithm}
\usepackage{array}
\usepackage{bbm}
\usepackage[caption=false,font=footnotesize,labelfont=footnotesize,textfont=footnotesize]{subfig}
\usepackage{textcomp}
\usepackage{stfloats}
\usepackage{url}
\usepackage{verbatim}
\usepackage{graphicx}
\usepackage{booktabs}
\usepackage{color}
\usepackage[table]{xcolor}
\usepackage{diagbox}
\usepackage{multirow}
\usepackage{cite}
\usepackage{flushend}
\usepackage{threeparttable}
\makeatletter
\let\NAT@parse\undefined
\makeatother
\usepackage{hyperref}
\usepackage{balance}
\hypersetup{pdfstartview=FitH,
            colorlinks=true,
            linkcolor=red,
            anchorcolor=blue,
            citecolor=green
            }

\newcounter{RNum}

\usepackage{soul}
\soulregister{\cite}7
\soulregister{\citep}7
\soulregister{\citet}7
\soulregister{\ref}7
\soulregister{\pageref}7
\IEEEoverridecommandlockouts                              

\definecolor{table_c}{RGB}{233,238,248}

\title{\LARGE \bf
Intelligent Fish Detection System with Similarity-Aware Transformer 
}

\author{Shengchen Li$^{1,\dagger}$, Haobo Zuo$^{2,\dagger}$, Changhong Fu$^{1,*}$, Zhiyong Wang$^{3}$, Zhiqiang Xu$^{3}$
\thanks{$^{1}$Shengchen Li and Changhong Fu are with the School of Mechanical Engineering, Tongji University, Shanghai 201804, China.
{\tt\small changhongfu@tongji.edu.cn}}
\thanks{$^{2}$Haobo Zuo is with the Department of Computer Science, University of Hong Kong, Hong Kong 999077, China.}
\thanks{$^{3}$Zhiyong Wang and Zhiqiang Xu are with the Fishery Machinery and Instrument Research Institute, Shanghai 200092, China.}
\thanks{${\dagger}$Equal contribution \hspace{0.3cm}*Corresponding author}
}

\begin{document}
\maketitle
\thispagestyle{empty}
\pagestyle{empty}

\begin{abstract}

Fish detection in water-land transfer has significantly contributed to the fishery.
However, manual fish detection in crowd-collaboration performs inefficiently and expensively, involving insufficient accuracy.
To further enhance the water-land transfer efficiency, improve detection accuracy, and reduce labor costs, this work designs a new type of lightweight and plug-and-play edge intelligent vision system to automatically conduct fast fish detection with high-speed camera. 
Moreover, a novel similarity-aware vision Transformer for fast fish detection (FishViT) is proposed to onboard identify every single fish in a dense and similar group.
Specifically, a novel similarity-aware multi-level encoder is developed to enhance multi-scale features in parallel, thereby yielding discriminative representations for varying-size fish.
Additionally, a new soft-threshold attention mechanism is introduced, which not only effectively eliminates background noise from images but also accurately recognizes both the edge details and overall features of different similar fish.
85 challenging video sequences with high framerate and high-resolution are collected to establish a benchmark from real fish water-land transfer scenarios. 
Exhaustive evaluation conducted with this challenging benchmark has proved the robustness and effectiveness of FishViT with over 80 FPS. Real work scenario tests validate the practicality of the proposed method. The code and demo video are available at \url{https://github.com/vision4robotics/FishViT}.
\end{abstract}

\section{Introduction}
\begin{figure}[t]
\centering
	{
	\includegraphics[width=1.0\linewidth]{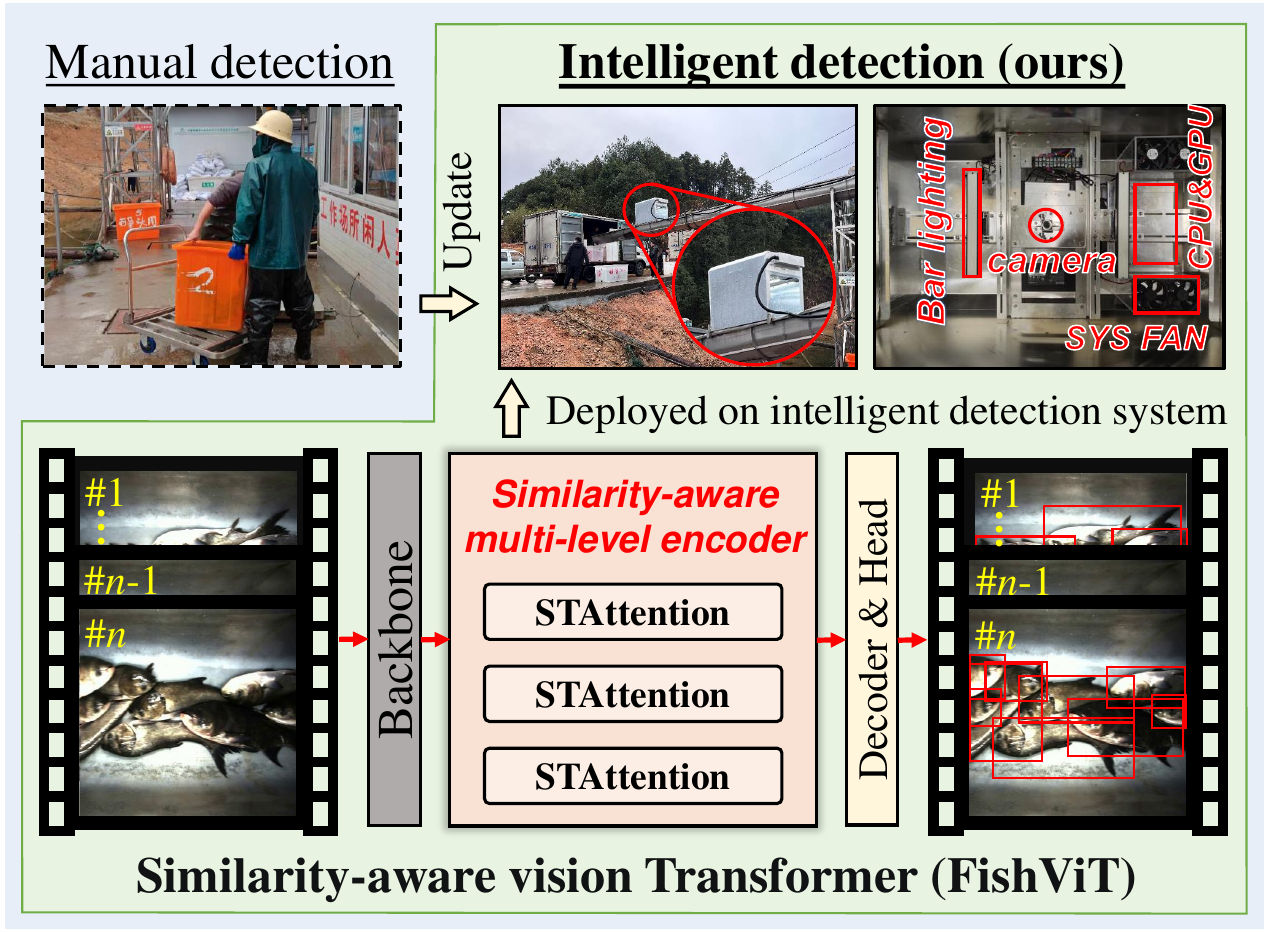}
	\centering
}
	\setlength{\abovecaptionskip}{-10pt}
	\caption
	{Fish detection in water-land transfer and the proposed fish detection method (FishViT). The picture with the dotted box represents the traditional manual detection mode, which is inefficient and costly. The proposed intelligent detection system significantly improves production efficiency, enhances accuracy, and reduces costs. Fresh fish slide fast from the pipeline into vehicles for sale and the intelligent detection device stand detects above the pipeline. The proposed FishViT is embedded into the intelligent detection system to effectively realize high-speed fish detection. FishViT mainly consists of three components: Backbone, Similarity-aware multi-level encoder, and Decoder \& Head. The Similarity-aware multi-level encoder is composed of three parallel soft-threshold attention (STAttention) modules.
	}
	\label{1}
	\vspace{-12pt}
\end{figure}

Intelligent vision systems can effectively solve the problems of low efficiency, low accuracy, and high cost associated with traditional crowd-collaboration mode for fish detection in water-land transfer.
Water-land transfer, \textit{i.e.}, transferring fresh fish from surface fish culture vessels to vehicles for sale, while keeping fish alive and injury-free.
Currently, as shown in Fig. \ref{1}, numerous terminals still rely on manual detection methods to detect fish during the process of water-land transfer.
However, due to the rapid expansion of the fishery market~\cite{zhong2020constructing}, traditional manual fish detection has gradually revealed its shortcomings of low efficiency, high cost, and low precision. The specific reasons mainly include the following aspects: \textit{1)} manual fish detection is prone to fatigue and errors due to the prolonged attention and concentration required; \textit{2)} this kind of detection is subject to inconsistencies that arise from variations in individual perception, training, and attention levels, leading to mix-ups and misidentifications; \textit{3)} this way relies heavily on human labor, which can be expensive, prone to fish injury, and challenging to scale up or down based on demand. 
\textbf{\textit{Therefore, how to design a high efficiency, high precision and fish-injury-free intelligent fish detection system is of great urgency.}}

With the development of deep learning, learning-based fish detection has drawn considerable attention due to its prosperous applications in fishery, \textit{e.g.}, fish economic benefit estimation
~\cite{thakur2023enhancing}, feedstuff feeding ~\cite{an2021application}, and culture density adjustment~\cite{wang2021intelligent}. It can bring enormous economic benefits to the fishery. 
Early learning-based detection methods are mostly based on convolutional neural networks (CNNs)~\cite{Girshick2015FastRC, Cai2018CascadeRC}.
However, since CNNs lack global information integration ability, CNN-based detection methods struggle to identify and locate the fish precisely in the presence of abundant highly similar appearance interference.

Recently, Transformer-based object detection methods have gained increasing popularity in the academic community due to their ability for global modeling, high robustness, and detection accuracy.~\cite{Li2022DN-DETR}.
Despite remarkable advancements, these techniques have difficulties to be applied for efficient and accurate fish detection. 
Firstly, the fish slide down at high speed in the pipeline for water-land transfer, which makes it hard to meet real-time application needs because of the high computational complexity of classical Transformer structures. 
Moreover, the presence of abundant highly similar fish in the water-land transfer pipeline and the frequent occurrence of challenging scenarios such as occlusion and cluttered backgrounds disturb the global information integration of Transformer.
\textbf{\textit{Thus, how to develop a lightweight, efficient, and robust Transformer-based detector to handle fish detection challenges, thereby satisfying the increasing demand for fish water-land transfer is an urgent problem.}}

The essential component of the Transformer is the attention mechanism. 
The performance of attention mechanisms has been greatly improved and innovated by state-of-the-art (SOTA) works~\cite{amjoud2023object}.
However, when facing fish detection in water-land transfer with high similarity and density, traditional attention shows poor robustness due to a lack of sufficient identification ability.

To address the aforementioned issues, the proposed similarity-aware vision Transformer for fast fish detection (FishViT) is deployed on the self-built intelligent vision device to effectively recognize every single fish in a dense and similar group, as indicated in Fig. \ref{1}.
Specifically, the similarity-aware multi-level encoder is designed to enhance multi-scale features in parallel, generating discriminative representations for fish of various sizes.
Importantly, the soft-threshold attention (STAttention) is introduced to suppress background noise thereby accurately recognizing the edge details of diverse, similar fish. 
The main contributions of this work are as follows:
\begin{itemize}

\item A lightweight and plug-and-play intelligent system is designed to automatically realize dense and fast fish detection with the onboard camera, heat sink, and processor. Compared with traditional manual detection, it can improve efficiency and accuracy at lower cost.

\item A similarity-aware vision Transformer for fast fish detection is introduced for the efficient detection of individual fish within dense and visually similar groups, deployed on the self-built intelligent system. A similarity-aware multi-level encoder is proposed to enhance multi-scale feature representation capabilities in parallel.

\item An innovative soft-threshold attention mechanism is presented to effectively eliminate background noise from images, thereby precisely discerning the edge information of diverse yet similar fish.  This mechanism aims to clarify every boundary of similar fish, leading to improved performance in fish detection.

\item Comprehensive evaluations on the 85 high-quality video sequences of real fish water-land transfer validate the promising performance of FishViT compared with other SOTA detectors. Work scenario tests have demonstrated the superior practicability of the proposed FishViT. 

\end{itemize}
\begin{figure*}[t]
  \centering
   \includegraphics[width=1.0\linewidth]{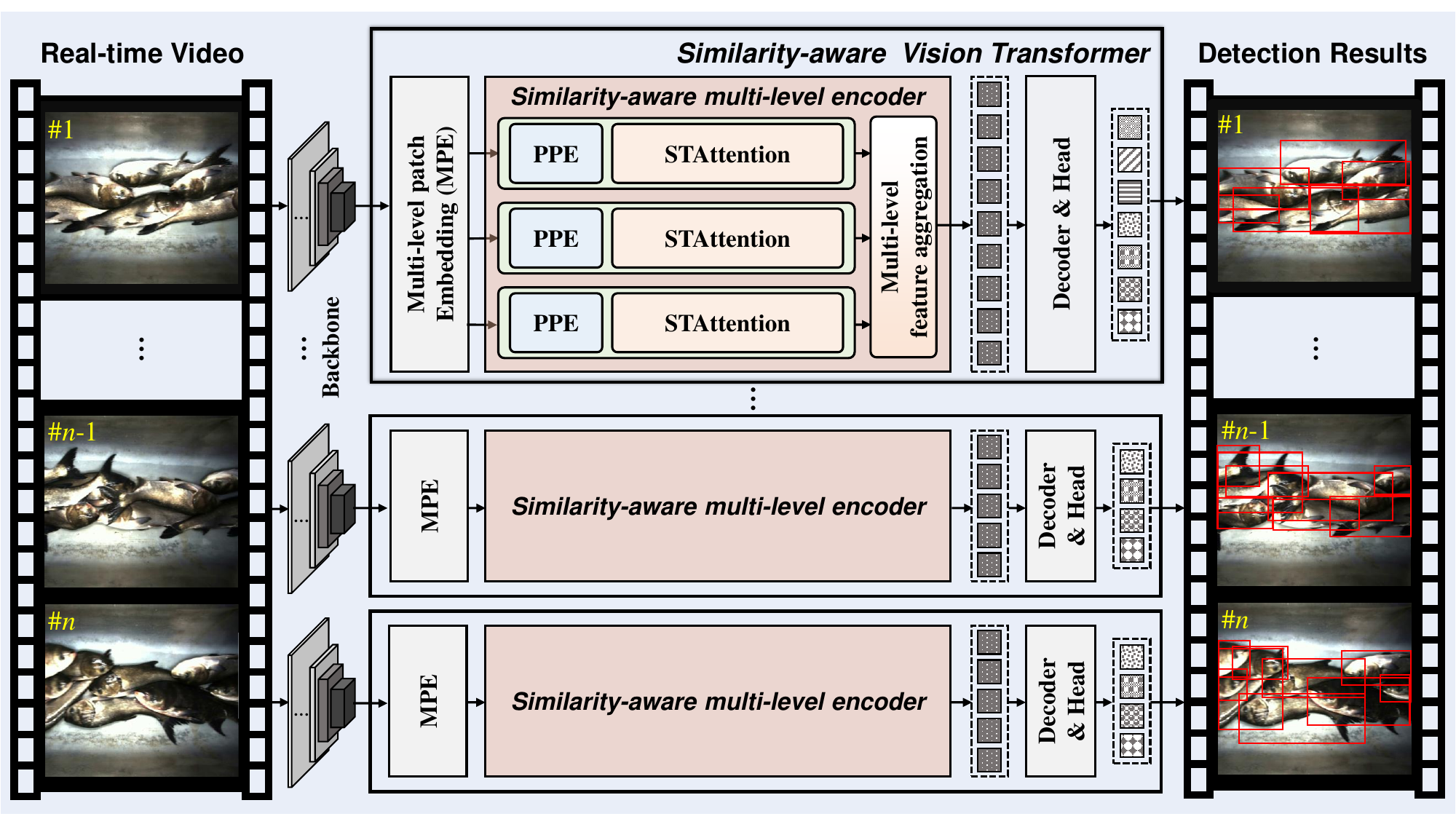}
   	\setlength{\abovecaptionskip}{-14pt}
   \caption{Overview of the proposed FishViT. The components from the left to right are \textit{Backbone}, \textit{Similarity-aware multi-level encoder}, \textit{Decoder \& Head}. The last three feature maps extracted by backbone are fed to similarity-aware multi-level encoder with parallel structure, each branch of encoder contains pooling positional encoding (PPE) and soft-threshold attention (STAttention). Finally, the feature map of each branch after multi-level aggregation are fed into decoder \& head for detection. Best viewed in color.}
   \centering
   \label{2}
   	\vspace{-10pt}
\end{figure*}

\section{Related Works}

\subsection{Fish Detection}
Compared with early manual fish detection, the rapid development of deep learning makes fish detection intelligent. Categorized by application scenarios, fish detection can be divided into two categories: underwater and overwater. 
Most modern fish detection applications and datasets focus on underwater scenarios and yield many impressive achievements~\cite{Park2019MarineVP}. 
Currently, there is little research has been done on detectors for fish water-land transfer.
In terms of detection methods, fish detectors can be categorized as CNN-based and Transformer-based detectors. CNN-based fish detectors, 
such as Faster R-CNN~\cite{Ren2015FasterRT} and 
YOLO~\cite{Knausgaard2022TemperateFD},
achieve promising results on various detection benchmarks. 
Compared with CNN-based detectors, the Transformer-based detectors have higher robustness and detection accuracy, 
such as DETRs~\cite{Carion2020End,Zhang2022DINODW}.
However, due to the high computational complexity of classical Transformer structures, the Transformer-based detectors are difficult to achieve a trade-off between high performance and real-time speed.
Furthermore, 
due to the presence of abundant highly similar fish in the water-land transfer pipeline and the frequent occurrence of challenging scenarios such as occlusion and cluttered backgrounds, 
existing fish detectors are difficult to apply effectively in water-land transfer.

\subsection{Attention Mechanism}
The attention mechanism plays a crucial role in deep learning with its outstanding capabilities for global feature modeling~\cite{Li2023MixingSA}. 
However, the high computational complexity of self-attention limits its application in visual tasks. There have been many research attempts to address this problem from multiple perspectives.
One line of research is using linear attention to address high computation complexity~\cite{Katharopoulos2020TransformersAR}, which replaces the Softmax function in self-attention with separate kernel functions.
K. M. Choromanski \textit{et} \textit{al.}~\cite{Choromanski2020RethinkingAW} propose Performers, approximating the Softmax operation with orthogonal random features. 
EfficientViT~\cite{Cai2023EfficientViTLM} uses depth-wise convolution to improve linear attention’s local feature extraction capacity.
W. Xu \textit{et} \textit{al.}~\cite{Xu2021Co} build on the above work to design a factorized attention mechanism, significantly enhancing computation efficiency. 
On the other hand, in order to further improve the model performance, M. Zhao \textit{et} \textit{al.}~\cite{Zhao2019DeepRS} add the soft-threshold ~\cite{Donoho1995NoisingBS} to deep residual networks, which effectively helps the model to better capture the key features and filter the noise. 
However, most of the above works are based on generalized scenario datasets, which are less robust when facing dense, occluded complex scenarios with highly similar fish in water-land transfer. 

\subsection{Feature Fusion}
Except for attention, another major challenge in object detection is effectively utilizing multi-scale features, which have been demonstrated to significantly improve performance, especially for varying-size objects. 
In modern CNN-based detectors~\cite{Tan2020EfficientDetSA}, 
feature pyramid network (FPN)~\cite{Lin2017FeaturePN} has become the primary solutions to exploit multi-scale features. 
Analogously, many Transformer-based detectors also attempt to improve DETRs by feature fusion. 
Deformable-DETR~\cite{Zhu2020DeformableDD} first introduces multi-scale features into DETR, exchanging information among multi-scale feature maps while improving the performance and convergence speed. 
SMCA-DETR~\cite{Gao2021FastCO} achieves efficient multi-scale feature coding in the encoder by introducing multi-scale self-attention and utilizes concatenate for feature fusion. 
CF-DETR~\cite{Cao2022CF-DETRCT} integrates the Transformer encoder within an FPN architecture to generate feature pyramids.
Iterative multi-scale feature aggregation~\cite{Zhang2023TowardsEU} designs a sparse sampling strategy for multi-scale  features to boost the performance of Transformer-based object detectors significantly with only slight computational overhead. 
Although the above methods enable the use of multi-scale features in Transformer-based detectors, they introduce huge computational overhead, making it difficult to effectively realize end-to-end real-time object detection.

\section{Methodology}

The workflow of FishViT is shown in Fig. \ref{2}. It can be divided into three modules: \textit{Backbone}, \textit{Similarity-aware multi-level encoder}, \textit{Decoder \& Head}. The encoder mainly consists of two novel parts, \textit{i.e.}, similarity-aware multi-level parallel structure, and soft-threshold attention mechanism.

\subsection{Backbone}
To meet the real-time applications onboard the intelligent vision system, FishViT uses a lightweight backbone known as ResNet18~\cite{He2016DeepRL}, which is used to extract multi-scale features. Specifically, the last three output feature maps of the backbone are utilized as the input to the subsequent process.

\noindent\textit{\textbf{Remark 1:}}
Given an image $\mathcal{I}\in\mathbb R^{{W}_{0}\times {H}_{0}\times 3}$, the backbone network generates its feature maps. For descriptive purposes, the last three output feature maps, which are further fed to the Transformer encoder, are uniformly represented by $\mathcal{F}_{l}\in \mathbb R^{W\times H\times C}$  in the following introduction (\textit{C}, \textit{W}, \textit{H} represent the channel, width, and height of the feature maps respectively, and \textit{l} $ \in$ \{3, 4, 5\}).

\subsection{Similarity-aware Multi-level Encoder}
The prime challenge to water-land transfer fish detection is the high similarity of fish, which tends to result in missed and false detections in high-density and occlusion scenarios.
To cope with this issue, a lightweight similarity-aware multi-level encoder based on STAttention is designed.
Specifically, the multi-level parallel structure is designed to enhance the multi-scale feature representation for addressing the fish appearance similarity issues.
Through the multi-level patch embedding, feature maps of different levels are obtained and fed to each branch in parallel.
The self-attention mechanism is utilized in every branch for global modeling and establishing context relationships.

\noindent\textit{\textbf{Remark 2:}}
Through the design of the similarity-aware multi-level encoder, different levels of features are fully extracted.
The proposed STAttention performs global modeling and establishes context relationships at different levels, which greatly improves the robustness of the model and increases the detecting accuracy.

\subsubsection{Pooling positional encoding}

Positional encoding is a critical component for the self-attention mechanism to be able to understand and process sequence data. Inspired by~\cite{Xu2021Co}, the pooling positional encoding method (PPE) is proposed to speed up the model while guaranteeing its effect. As shown in Fig. \ref{4}, the process of PPE can be described by the formula as:
\begin{equation}
\begin{aligned}
  \mathcal{M}_{P} &=\mathrm{AvgPooling}(\mathcal {M})+\mathcal {M}\quad ,\\[1mm]
   \end{aligned}
\end{equation} 
where $\mathcal {M} $ represents the feature vector that after Multi-level patch embedding (MPE), and $\mathcal{M}_{P}$ represents the $\mathcal {M} $ completed the positional encoding.

\noindent\textit{\textbf{Remark 3:}}
The use of the proposed pooling position encoding effectively reduces redundant information and speeds up computation. $\mathcal{F}\in \mathbb R^{W\times H\times C}$ and $\mathcal{M} \in \mathbb R^{N\times C}$  represent the input and output of the MPE separately in the following introduction. 

\subsubsection{STAttention}

Given an input feature \(\mathcal{F} \in \mathbb{R}^{W \times H \times C}\), the representation is first transformed into the embedding space, denoted as \(\mathcal{M} \in \mathbb{R}^{N \times C}\), with a feature dimension of \(C\). The generalized attention function can be computed as~\cite{ qin2022cosformer}:
\begin{equation}
 \setlength{\abovedisplayskip}{10pt}
    \begin{aligned}
     \mathrm{Att}(\mathcal{M}) &= \sum{\frac{\mathcal{S}(\mathbf{Q},\mathbf{K}) }{\sum{\mathcal{S}(\mathbf{Q},\mathbf{K})}} \mathbf{V}} \quad ,\\[1mm]
   \end{aligned}
\label{Eq:2}
\end{equation}
where \(\mathcal{S}(\cdot)\) denotes the function for measuring the similarity between queries and keys, if \(\mathcal{S}(\mathbf{Q}, \mathbf{K}) = \exp(\mathbf{Q}\mathbf{K}^{\mathrm{T}})\), then Eq. \ref{Eq:2} simplifies to the $\mathit{scaled \quad dot-product}$  attention mechanism with softmax normalization. However, this approach incurs a computational complexity of \(O(N^2C)\), rendering it unsuitable for real-time applications.
The use of decomposable similarity functions enables attentions to be computed in a linear manner. Specifically, two functions \(\phi(\cdot)\) and \(\psi(\cdot)\) are used and the second matrix multiplication is computed:
\begin{equation}
 \setlength{\abovedisplayskip}{10pt}
    \begin{aligned}
     \mathcal{S}(\mathbf{Q},\mathbf{K})\mathbf{V}&=(\phi(\mathbf{Q}) \psi (\mathbf{K})^{T}) \mathbf{V}=\phi(\mathbf{Q})(\psi(\mathbf{K})^{T}\mathbf{V}) \quad ,\\[1mm]  
   \end{aligned}
\end{equation}
the result leads to a \(O(NC^2)\) computation complexity, which greatly reduces its complexity in comparison for \textit{N} is generally much larger than \textit{C} value for images.
Factorized attention (FactorAtt) is obtained when $\phi$ is the identity function and $\psi$ is the softmax~\cite{Xu2021Co}:
\begin{equation}
    \begin{aligned}
     \mathrm{FactorAtt}(\mathbf{X}) &= \frac{\mathbf{Q}} {\sqrt{C}} (\mathrm{Softmax} (\mathbf{K})^{T}\mathbf{V}) \quad .\\[1mm]
   \end{aligned}
\end{equation}

\begin{figure}[t]
  \centering{
   \includegraphics[width=0.99\linewidth]{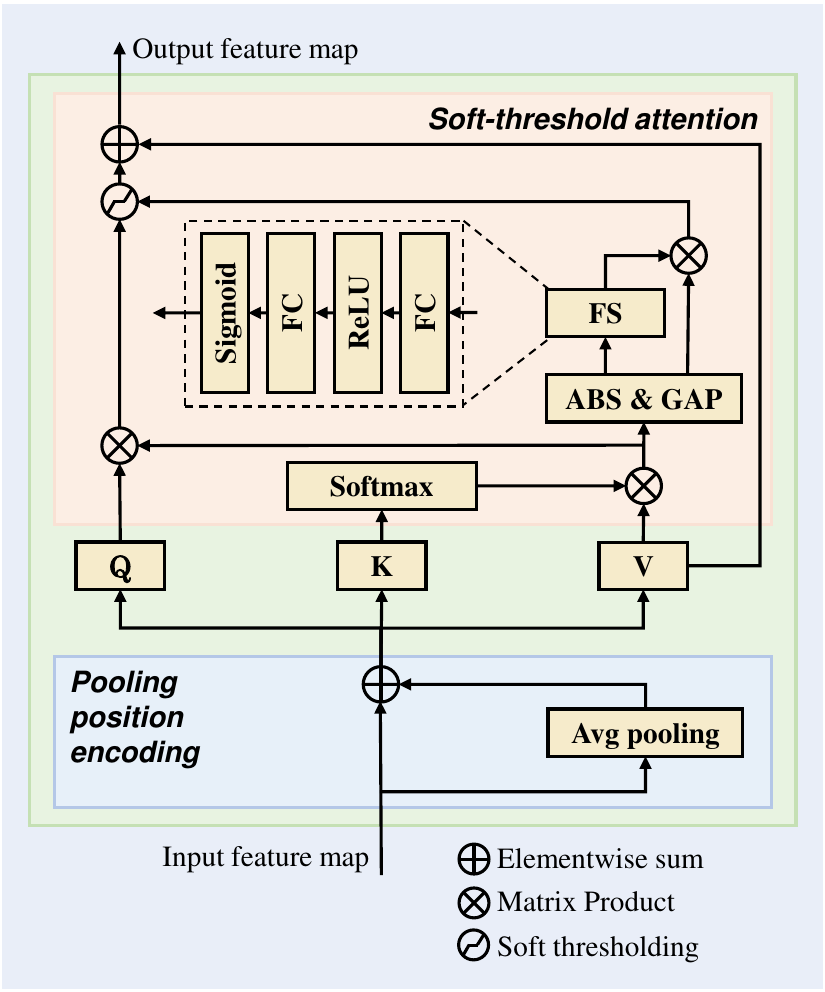}
   	\centering
}
   		\setlength{\abovecaptionskip}{-14pt}
     
   \caption{Detailed workflow of the STAttention. The input feature maps are processed through STAttention after pooling position encoding.  The result of $\mathrm{Softmax}(\mathbf{K})^{T}\mathbf{V}$ is used to generate a soft-threshold, and the linear attention result is filtered by the soft-threshold. Specifically, the soft-threshold mechanism consists of two main modules: ABS \& GAP and FS. ABS stands for Absolute Value, and GAP represents Global Average Pooling, while the FS module refers to the operations within the dashed box. Finally, $\mathbf{V}$ is added to the filtered result through the shortcut technique, and the added result is used as the final output feature map. This soft-threshold mechanism effectively suppresses background noise, enabling precise identification of the edges of each individual fish within within dense and visually similar groups. Best viewed in color.}
   \label{4}
   		\vspace{-10pt}
\end{figure}

In the face of high speed and high similarity dense fish detecting scenarios, the means of denoising can convert useful information into active and effective features while turning noisy information into invalid or near-zero features, which enables the detector to better discriminate the edge information of the fish and improve the detection accuracy.
The combination of soft-threshold (ST) and deep learning is a promising method for denoising~\cite{isogawa2017deep}, and soft-threshold calculation formula can be expressed as follows: 
\begin{equation}
\begin{aligned}
\mathrm{ST}(x)= \left \{
\begin{array}{cc}
x-\tau  \quad , & x > \tau\\
\quad 0 \quad\quad , & -\tau < x < \tau\\
x+\tau  \quad , & x < - \tau
\end{array}
\right.,
\end{aligned}
\end{equation}
where $\mathit{x}$ is the input feature, $\tau$ indicates the threshold.
Inspired by~\cite{Zhao2019DeepRS}, a deep residual shrinkage network is introduced to automatically determine the threshold. Firstly, as shown in Fig. \ref{4},  the absolute operation and GAP layer are used for the result of $(\mathbf{K})^{T}\mathbf{V}$ to simplify the feature mapping  to 1-D vector, and then the feature dimension is reduced through FS module.
Eventually, the output feature of FS is scaled to the range (0,1) by sigmoid layer, the formula is described as:
\begin{equation}
    \begin{aligned}
     \alpha _{c} =\frac{1}{1+e^{-z_{c} } }\quad ,\\[1mm]
   \end{aligned}
\end{equation}
where $\alpha _{c}$  is the scaling parameter of feature map $\mathit{cth}$ channel, $\mathit{z}_{c}$ is the feature at the $\mathit{c th}$ neuron of 1-D vector.
Ultimately, the thresholds are calculated by the formula as:
\begin{equation}
\tau _{c} =\alpha _{c} \cdot \mathrm{average}\left | ((\mathbf{K})^{T}\mathbf{V})_{i,j,c}  \right | \quad ,\\[1mm]
\label{Eq:7}
\end{equation}
 where  $\tau _{c}$  is the threshold for the $\mathit{cth}$ channel of the feature map, and \textit{i}, \textit{j}, and \textit{c} are the indexes of width, height, and channel of the feature map, respectively.

Meanwhile, the shortcut technique is introduced to solve the problem of gradient vanishing or gradient explosion in deep neural network training.
In summary, the formula given in STAttention is as follows:
\begin{equation}
 \setlength{\abovedisplayskip}{10pt}
    \begin{aligned}
     \mathrm{STAttention}(\mathbf{X}) &= \mathrm{ST}(\frac{\mathbf{Q}} {\sqrt{C}} (\mathrm{Softmax} (\mathbf{K})^{T}\mathbf{V})) +  \mathbf{V} \quad .\\[1mm]
   \end{aligned}
\setlength{\belowdisplayskip}{10pt}
\end{equation}

\noindent\textit{\textbf{Remark 4:}}
STAttention accelerates the model speed by linearization of self-attention. Meanwhile, through the setting of soft-threshold, the influence of background noise can be better eliminated, thus precisely discerning the edge information of diverse yet similar fish, improving the detection accuracy.
The specific process of Eq. \ref{Eq:7} is:
\begin{equation}
 \setlength{\abovedisplayskip}{10pt}
    \begin{aligned}
     \mathrm{STAttention}(\mathbf{X}) &= \mathrm{ST}(\frac{\mathbf{Q}} {\sqrt{C}} (\mathrm{Softmax} (\mathbf{K})^{T}\mathbf{V})) +  \mathbf{V} 
     \\ &= \mathrm{Sign}( \mathrm{FactorAtt}(\mathbf{X})) \tau _{c}  +  \mathbf{V} \quad ,\\[1mm]
   \end{aligned}
\setlength{\belowdisplayskip}{10pt}
\end{equation}
where $\mathrm{Sign(*)}$ represents the sign function.
\begin{table*}[t]
\vspace{-3pt}
\renewcommand{\arraystretch}{1}
\caption{Main detection results. \textcolor{red}{Red} represents the best result and \textcolor{blue}{blue} represents the second best result. $\uparrow$ indicates that a higher value for this metric is better, while $\downarrow$ signifies that a lower value is preferable for this metric. 
All real-time detectors share a common input size of 640 and all end-to-end object detectors share a common input  size of 800. Our FishViT achieved the best performance across all metrics.}
\vspace{-3pt}
\label{tab:parametric}
\centering
\colorbox{table_c}{
 \begin{tabular}{m{4.5cm} m{1.2cm}<{\centering} m{1.2cm}<{\centering} m{1.2cm}<{\centering} m{1.1cm}<{\centering} m{1.1cm}<{\centering} m{1.1cm}<{\centering} m{1.1cm}<{\centering} m{1.1cm}<{\centering}}
 \toprule
 $\textbf{Methods}$ & $\textbf{Backbone}$ & $\textbf{GFLOPs}\downarrow$ &$ \textbf{Params}\downarrow$ & $\textbf{FPS}_{bs=16}\uparrow$& $\textbf{AP}_{50}^{val}\uparrow$ & $\textbf{E}_{cls}\downarrow$ & $\textbf{E}_{loc}\downarrow$& $\textbf{E}_{miss}\downarrow$\\ 
 \hline
 \hline
\multicolumn{9}{l}{\cellcolor{black!20} \textit{Real-time End-to-end Object Detectors}}\\
 YOLOv8-M &  & 79 & 26 M & 62.6 & 93.2 & 0.48 & 3.29 & 1.91\\
 YOLOv8-L &  & 165 & 79 M & 30.3 & 93.6 & 0.27 & 2.85 &  1.47\\
 YOLOv8-X &  & 258 & 165 M & 17.7 & \textcolor{blue}{94.4} & \textcolor{blue}{0.24} & 2.6 &  1.23\\ 
\hline
 \hline
 \multicolumn{9}{l}{\cellcolor{black!20} \textit{End-to-end Object Detectors}}\\
 DETR-DC5   &	R50 & 187 & 41 M & 5.1 & 84.1 & 13.38 & 10.92 &  2.36\\
 DETR-DC5	&   R101 & 253 & 60 M & 3.7 & 84.5 & 16.31 & 16.74 &  3.96\\
 Anchor-DETR-DC5	&  R50 & 172 & 39 M & 5.4 & 87.6 & 2.46 & 1.53 & 0.38\\
 Anchor-DETR-DC5	&  R101 & 240 & 58 M & 1.5 & 89.6 & 2.71 & 1.49 & \textcolor{blue}{0.36}\\
 Conditional-DETR-DC5  &	R50 & 195 & 44 M & 4.3 & 91.1 & 2.59 & 2.14 & 0.5\\
 Conditional-DETR-DC5  &	R101 & 262 & 63 M & 2.6 & 92.7 & 4.37 & 6.78 & 1.23 \\
\hline
 \hline
 \multicolumn{9}{l}{\cellcolor{black!20} \textit{Real-time End-to-end Object Detectors}}\\
 RT-DETR	&  R18 & \textcolor{blue}{56.9} & \textcolor{blue}{20 M} & \textcolor{blue}{64.7} & 93.6 & 0.29 & \textcolor{blue}{1.85} & 0.61\\
 \textbf{FishViT (ours)}	&  R18 & \textbf{\textcolor{red}{45.6}} & \textbf{\textcolor{red}{18 M}} & \textbf{\textcolor{red}{82.3}} & \textbf{\textcolor{red}{94.7}} & \textbf{\textcolor{red}{0.11}} & \textbf{\textcolor{red}{1.65}} &\textbf{\textcolor{red}{0.28}} \\
 \bottomrule
 \end{tabular}}
 \vspace{-13pt}
 \end{table*}
\subsection{Decoder \& Head}
As shown in Fig. \ref{2}, the similarity-aware multi-level encoder transforms multi-scale features into a sequence of image features. 
In order to provide more encoder features with accurate classification and precise location for object queries, the output sequence of the encoder needs to go through the IoU-aware query selection~\cite{Lv2023DETRsBY} to obtain a fixed number of image features as initial object queries for the decoder.  
Then, the selected object queries are optimized by the decoder and mapped to classification scores and bounding boxes by the prediction head.
Finally, the proposed FishViT uses a Transformer decoder with auxiliary prediction heads to get fish detection results. 

\begin{figure}[t]
\centering
	{
	\includegraphics[width=1.0\linewidth]{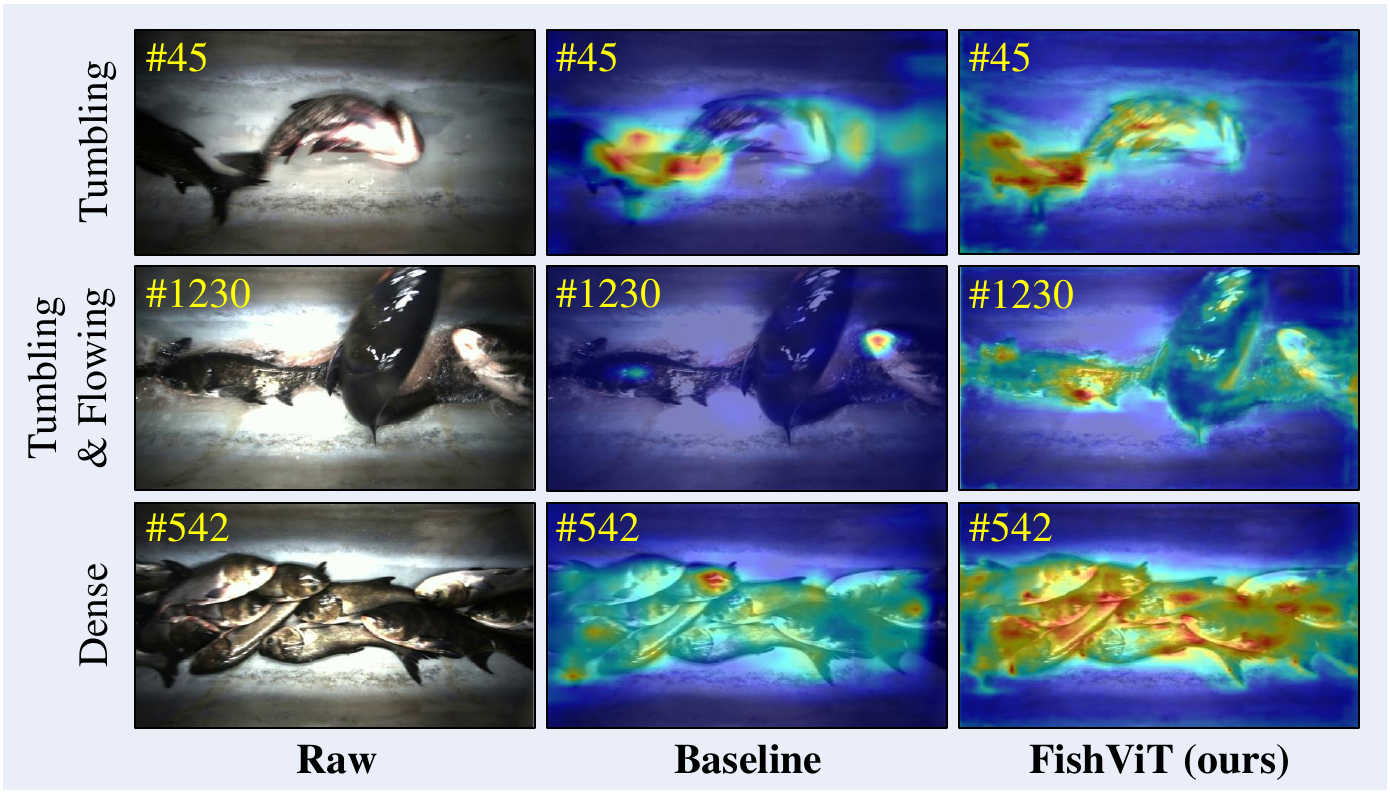}
	\centering
}
	\setlength{\abovecaptionskip}{-14pt}
	\caption
	{ Visualization of the confidence maps of the Baseline and the proposed FishViT. FishViT can effectively reduce background interference and focus on the detailed representation of the
fish to cope with extremely complex situation such as high-speed tumbling, flowing water, and high density. Best viewed in color.
}
	\label{3}
	\vspace{-10pt}
\end{figure}

\noindent\textit{\textbf{Remark 5:}}
As shown in Fig.  \ref{3}, in challenging scenarios such as tumbling, flowing, and dense, FishViT is able to clearly discriminate the edges of each fish in a dense fish group by benefiting from the parallel multi-scale representation as well as the denoising effect of STAttention.
\section{Experiments}
\subsection{Implementation Details and New Benchmark}
This work develops a new type of lightweight plug-and-play intelligent vision system to implement high-speed dense fish detection autonomously.
The whole system mainly includes the following two components: a pipeline with a total length of 16 meters and a retractable tail to match different types of land sales vehicles.
An intelligent detection device with a high framerate camera (to capture high-speed images of fish), a high-performance processor (Intel \textit{i}9-12900KF+NVIDIA RTX 3060), two bar lightings (to avoid outdoor light interference), four SYS FAN (for heat dissipation), a metal housing (for light resistant and waterproof), and an indoor console. 
Utilizing the proposed intelligent vision system, 85 challenging video sequences with a high frame rate of 90 frames/s are collected in real fish water-land transfer scenarios to comprehensively assess FishViT's effectiveness in fish detection.
The video sequences contain two fish species, \textit{i.e.}: 'Black carp' and 'Silver carp', of which 50 sequences are used for training, 5 sequences are used for validation and 30 sequences for testing.
The sequences share six challenging attributes, including 'spume', 'overexposure', 'tumbling', 'dense', 'flowing', and 'underexposure'.
ResNet18~\cite{He2016DeepRL} serves as the backbone for FishViT. 
FishViT is trained by AdamW optimizer with the values of $weight\underline{\enspace}decay$ and $base\underline{\enspace}learning\underline{\enspace}rate$ as $10^{-4}$ for 150 epochs.
\begin{figure*}[t]
\centering
	{
	\includegraphics[width=1.0\linewidth]{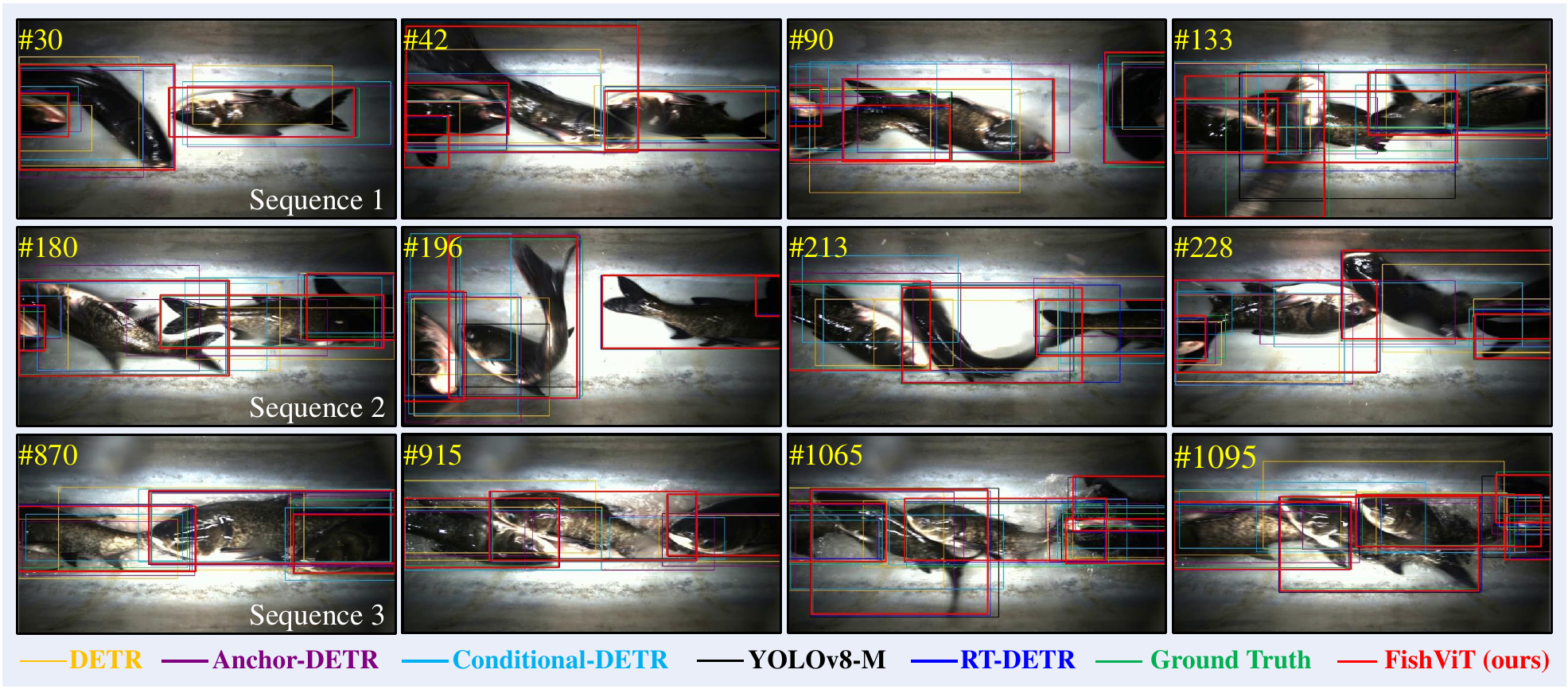}
	\centering
}
	\setlength{\abovecaptionskip}{-14pt}
	\caption
	{Comparison of FishViT results with other SOTA detectors. YOLOv8-M has duplicate detection boxes due to post-processing, while other Transformer-based detectors exhibit missed detections. Our FishViT achieved the best results. Best viewed in color.
}
	\label{detect_result}
	\vspace{-10pt}
\end{figure*}
\subsection{Evaluation on Detection Benchmarks}
To fully assess FishViT's fish detection capabilities, the crucial detection accuracy must be evaluated. 
The evaluation metric used is the standard COCO AP metric with a single scale image as input~\cite{Lin2014COCO}. 
In object detection tasks, Average Precision (AP) is usually used as an indicator to measure the performance of a model. 
However, in different application scenarios, it is not appropriate to just look at the value of AP~\cite{bolya2020tide}. 
In the water-land fish transfer scenario, classification error $(\textbf{E}_{cls})$ is intolerable because it directly relates to the selling price. 
In addition, detection tasks are often performed to serve downstream tasks, such as fish counting and fish size recording. 
Localization error $(\textbf{E}_{loc})$, and missed GT error $(\textbf{E}_{miss})$ can easily degrade the performance of downstream tasks. 
Therefore, the above three error types need to be comprehensively evaluated.

Real-time detectors have received widespread attention and applications due to their excellent performance as well as superior efficiency. 
YOLOs are the cutting-edge CNN-based real-time detectors, thereby YOLOv8 ~\cite{Jocher_Ultralytics_YOLO_2023} with SOTA performance is selected as a representative for comparative experiments.
DETRs are Transformer-based cutting-edge end-to-end detection method, and representative DETR~\cite{Carion2020End}, Anchor-DETR~\cite{wang2022anchor}, Conditional-DETR~\cite{meng2021conditional}, and RT-DETR~\cite{Lv2023DETRsBY} are selected for comparison experiments.
For fairness, all real-time detectors adopted the same training strategy, and end-to-end object detectors are only fine-tuned according to official recommendations. 
TABLE~\ref{tab:parametric} shows the overall detection performance. 
FishViT achieves a maximum FPS of \textbf{82.3}, which is fully meets the real-time requirements for fish detection in water-land transfer.
Attributed to the excellent multi-scale feature expression ability of the similarity-aware multi-level encoder and the high similarity-aware capability of STAttention, FishViT yields the best $\textbf{\text{AP}}_{50}$  $(\textbf {94.7})$. 
Meanwhile, FishViT reaches a minimum of \textbf{0.11} on the most concerned $\textbf{E}_{cls}$, and a minimum of \textbf{1.65} and \textbf{0.28} on the $\textbf{E}_{loc}$ and $\textbf{E}_{miss}$, respectively. 
Figure \ref{detect_result} visualizes the results of FishViT and of other SOTA detectors, with FishViT demonstrating superior performance. 

\subsection{Evaluation on Attributes}
\begin{figure}[t]
  \centering
   \includegraphics[width=0.99\linewidth]{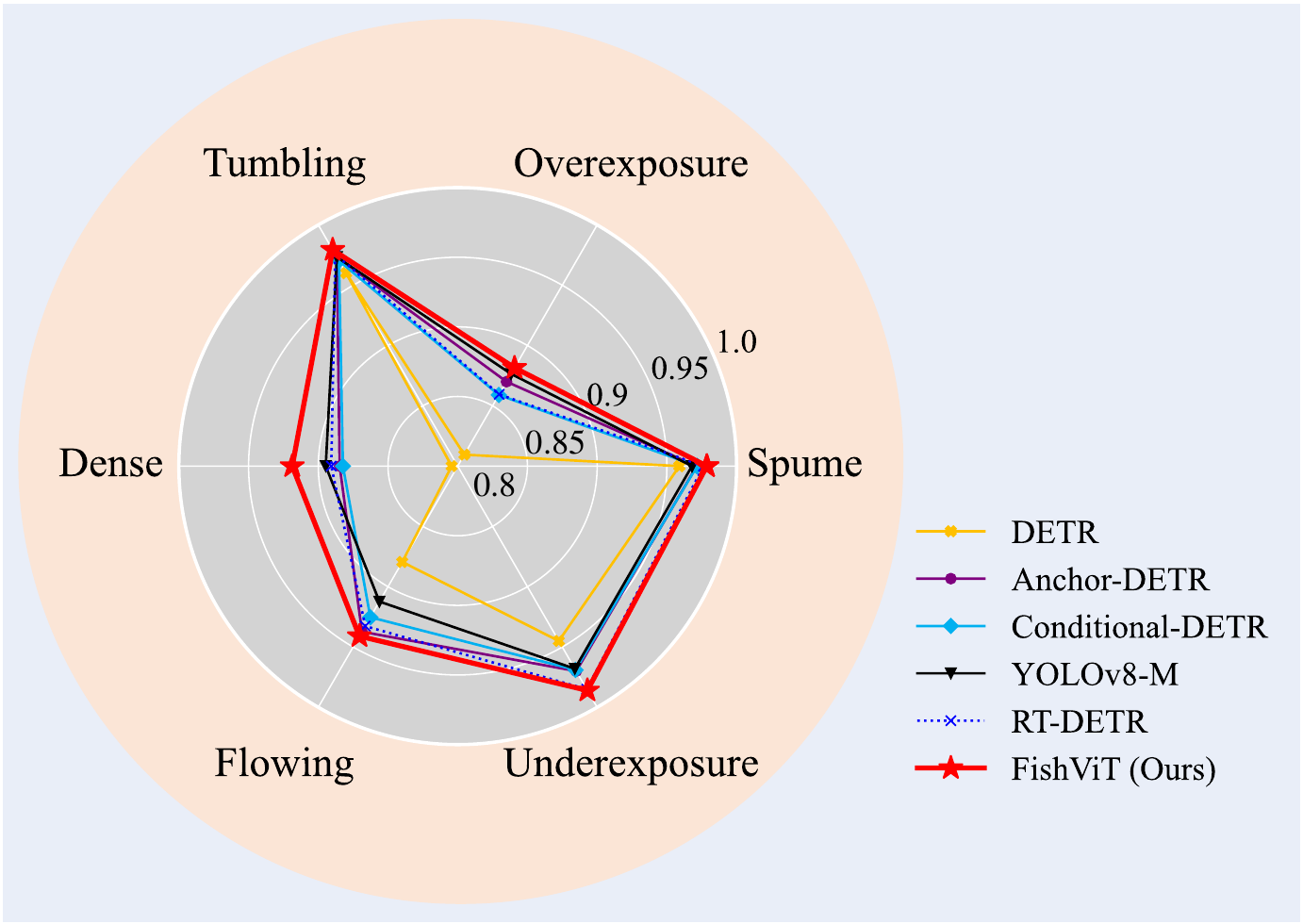}
   		\setlength{\abovecaptionskip}{-14pt}
   \caption{The detection accuracy of FishViT and other SOTA detectors under six challenging attributes. Our FishViT achieves the best performance across all six challenging attributes. Best viewed in color.}
   \centering
   \label{5}
   		\vspace{-10pt}
\end{figure}

To exhaustively evaluate the performance of FishViT in fish detection, attribute-based detection accuracy evaluation experiments are conducted based on 30 sequences used for test. 
$\textbf{AP}_{50}$ is used as the evaluation metric of the selected videos for each challenging attribute.
As shown in Fig.~\ref{5}, FishViT can consistently achieve satisfactory performance with optimal detection accuracy in all six challenging scenarios. 
Compared to other attributes, FishViT's benefits are most evident in the challenging scenario of dense. 
This is the primary factor considered in FishViT design, \textit{i.e.}, to discriminate every single fish in the dense group with highly similar appearances.
STAttention mechanism can effectively eliminate background noise from images, while precisely discerning the edge information of diverse yet similar fish. 
This mechanism aims to clarify every boundary of similar fish and, as a result, improves detection accuracy in challenging scenarios.
Furthermore, multi-scale feature representation capability is enhanced by introducing similarity-aware multi-level encoder with parallel structure, which further improves the detection accuracy of similar fish.

\begin{table}[b]
	\centering
	\caption{Ablation study of various parts of the proposed FishViT. 
 $\Delta$ symbolizes the improvement
    over the Baseline method.}
    \label{table_abl_study}
	\resizebox{\linewidth}{!}{
        \colorbox{table_c}{
		\begin{tabular}{lcccccc|c} 
		 \hline
			Detecting Methods & $\textbf{AP}_{50}$   & $\Delta_{{AP}_{50}}$ & $\textbf{E}_{cls}$ & $\Delta_{E_{cls}}$ \\
		\hline 
			Baseline  & 89.1  & -  & 1.42 & -  \\ 
			Baseline+ST & 92.6 & +3.5 &  0.30  & -1.12\\
			Baseline+ML & 93.2 & +4.1 & 0.23 & -1.19\\
		\hline 
  $\textbf{Baseline}$+$\textbf{ST}$+$\textbf{ML}$ ($\textbf{FishViT}$) & $\textbf{94.7}$ & $\textbf{+5.6}$ & $\textbf{0.11}$ & $\textbf{-1.31}$ \\
		 \hline 
	     \end{tabular}}}
\end{table}

\subsection{Ablation Study}
To verify the effectiveness of each module in the proposed method, FishViT with different modules enabled is studied.
This work considers Baseline as the model with only factorized attention encoder~\cite{Xu2021Co}.
ST represents the STAttention and ML represents the parallel structure in similarity-aware multi-level encoder.
The most important error in this fish detection task is the classification error, as it is directly linked to the revenue.
Therefore, the ablation experiments used only $\textbf{\text{AP}}_{50}$ and classification error as evaluation metrics.

\subsubsection*{\bf Discussion on STAttention}
As shown in TABLE~\ref{table_abl_study}, with the addition of the soft-threshold (Baseline+ST), $\textbf{\text{AP}}_{50}$ directly increased by \textbf{3.5}, indicating that STAttention can effectively improve the accuracy of the model.
In addition, the classification error directly decreased by \textbf{1.12}, indicating that STAttention is able to remove the noise interference to a great extent and improve the classification accuracy under the condition of highly similar fish.
\subsubsection*{\bf Discussion on Similarity-aware Multi-level Encoder}
As shown in TABLE~\ref{table_abl_study}, adding two extra Baseline branches (Baseline+ML) increases $\textbf{\text{AP}}_{50}$ by \textbf{4.1}, demonstrating the multi-branch structure's positive impact on detection accuracy. Additionally, the classification error decreases by \textbf{1.19}, suggesting that the parallel structures help the model better capture edge information and improve classification accuracy by integrating multi-scale semantic data.

By combining ST and ML, FishViT (Baseline+ST+ML) efficiently learns the edge information while further enhancing the fish edge discrimination through the noise reduction effect of STAttention, which makes the $\textbf{\text{AP}}_{50}$ rises by \textbf{5.6} and reducing classification error by \textbf{1.31}, fully demonstrating the strong robustness and high accuracy of FishViT.
\begin{figure}[tbp]
	\centering
	\includegraphics[width=1.0\linewidth]{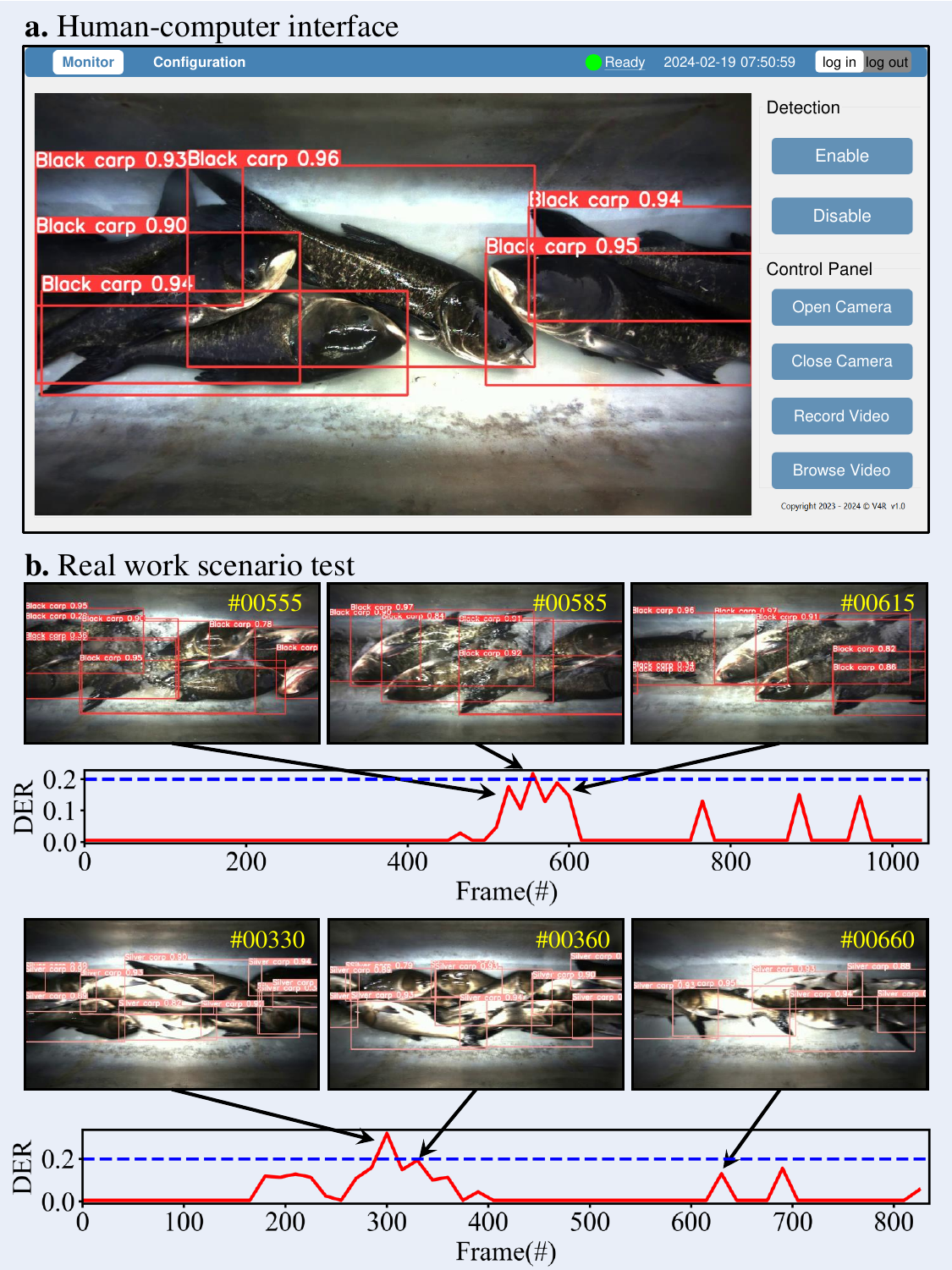}
	\caption{Visualization of the real-world scenario test. \textbf{a.} The specially designed fish detection software interface. \textbf{b.} The detection results are marked with \textcolor{red}{red} boxes. The detection error rate (DER) is defined as \(1 - \text{AP}_{50}\) per frame. A DER score below the \textcolor{blue}{blue} dotted line is considered accurate and reliable detection in the real-world test. Best viewed in color.}
	\label{realworld_test}
\end{figure}
\subsection{Real Work Scenario Test}
The practicability of FishViT is further validated in real water-land transfer work scenario.  
FishViT achieves an average speed of over 80 FPS during the test, meeting real-time requirements.
The unique software interface for fish detection and two real-world test sequences are shown in Fig.~\ref{realworld_test}. 
Real-time images will be displayed in the interface's main window. 
A series of functions such as detection and recording can be achieved by clicking the button on the right, which is simple but efficient.
The two tests contain several typical fish detection challenges, including dense and flowing.
Test 1 demonstrates that high accuracy detection is maintained even as fish flipping and water velocity affect detection performance.
In Test 2, FishViT encounters challenging, complex scenes.
Nevertheless, FishViT manages to get remarkably elevated detection accuracy, which can be ascribed to STAttention's outstanding ability to identify similar objects.
In conclusion, FishViT can accurately identify every fish in complex and challenging scenarios, which is incredibly convenient for real-world fish water-land transfer.
\section{Conclusion}
A new type of lightweight, low-power, environmentally independent, and high-speed intelligent detection system deployed with the proposed FishViT is designed to automatically conduct high-speed fish detection in this work.
The objective is to solve the problem of low efficiency and high cost of fish detection in traditional crowd-collaborative water-land transfer mode. 
To cope with the high similarity, high speed, and high density problems of fish water-land transfer, a novel real-time end-to-end detector for fish detection is proposed. Additionally, STAttention with soft-threshold and similarity-aware multi-level encoder with parallel structure are presented. 
Extensive experiments prove that FishViT is capable of detecting fish at high speeds while delivering outstanding performance.
FishViT will significantly enhance subsequent critical tasks in the field of water-land transfer, including the tracking, segmentation, sizing, and counting of fish, providing robust academic support for these processes.
In conclusion, we firmly believe that the intelligent vision system with FishViT can aid in the advancement of fish detection in water-land transfer.

\section*{Acknowledgments}
This work is supported by Hangzhou Qiandao Lake Development Group Co. LTD and Fishery Machinery and Instrument Research Institute, Chinese Academy of Fishery Sciences.

\balance
\bibliographystyle{IEEEtran}
\bibliography{ref}
\end{document}